\documentclass[conference]{IEEEtran}
\IEEEoverridecommandlockouts
\usepackage{cite}
\usepackage{amsmath,amssymb,amsfonts}
\usepackage{algorithmic}
\usepackage{graphicx}
\usepackage{textcomp}
\usepackage{xcolor}
\usepackage{booktabs}
\usepackage{tabularx}
\usepackage{authblk}

\def\BibTeX{{\rm B\kern-.05em{\sc i\kern-.025em b}\kern-.08em
    T\kern-.1667em\lower.7ex\hbox{E}\kern-.125emX}}
\begin{document}

\title{Divide-and-Conquer: Dual-Hierarchical Optimization for Semantic 4D Gaussian Spatting}

\author[1,2,*]{Zhiying Yan\thanks{\textsuperscript{*}These authors contributed equally.}}
\author[1,2,*]{Yiyuan Liang}
\author[3]{Shilv Cai}
\author[1,2]{Tao Zhang}
\author[1,2]{Sheng Zhong}
\author[1,2]{Luxin Yan}
\author[1,2,$\dagger$]{Xu Zou\thanks{\textsuperscript{$\dagger$}Corresponding Author.}}

\affil[1]{Huazhong University of Science and Technology, China}
\affil[2]{National Key Laboratory of Multispectral Information Intelligent Processing Technology, China}
\affil[3]{Nanyang Technological University, Singapore}

\maketitle

\begin{abstract}
Semantic 4D Gaussians can be used for reconstructing and understanding dynamic scenes, with temporal variations than static scenes. Directly applying static methods to understand dynamic scenes will fail to capture the temporal features. Few works focus on dynamic scene understanding based on Gaussian Splatting, since once the same update strategy is employed for both dynamic and static parts, regardless of the distinction and interaction between Gaussians, significant artifacts and noise appear. We propose Dual-Hierarchical Optimization (DHO), which consists of Hierarchical Gaussian Flow and Hierarchical Gaussian Guidance in a divide-and-conquer manner. The former implements effective division of static and dynamic rendering and features. The latter helps to mitigate the issue of dynamic foreground rendering distortion in textured complex scenes. Extensive experiments show that our method consistently outperforms the baselines on both synthetic and real-world datasets, and supports various downstream tasks. Project Page: https://sweety-yan.github.io/DHO.
\end{abstract}


\section{Introduction}
\label{sec:intro}

Novel view synthesis and scene understanding play crucial roles in many applications such as autonomous driving \cite{zhou2024drivinggaussian}, VR/AR \cite{xie2024physgaussian}, and filmmaking. Combining both capabilities will further enhance spatial intelligence. With the advent of 3D Gaussian Splatting (3DGS) \cite{3dgs}, the speed of rendering has significantly increased, leading to rapid advancements in 3D/4D rendering, generation \cite{yi2024gaussiandreamer}, and understanding \cite{qin2024langsplat}. 

Most recent 3D scene reconstruction and understanding research \cite{qin2024langsplat} focuses on static scenes. While these methods demonstrate impressive performance on static scenes, they are ill-equipped to handle dynamic scenes due to the inability of vanilla 3DGS to capture the time-varying geometry and features of dynamic foreground objects. Recent works \cite{wu20244d} elevate 3D scenes to 4D, enabling high-fidelity rendering of dynamic scenes. Nonetheless, these approaches lack semantic-level understanding of dynamic scenes and struggle to support various 4D semantic downstream tasks. 
\begin{figure}[t]
\centerline{\includegraphics[width=0.98\linewidth]
{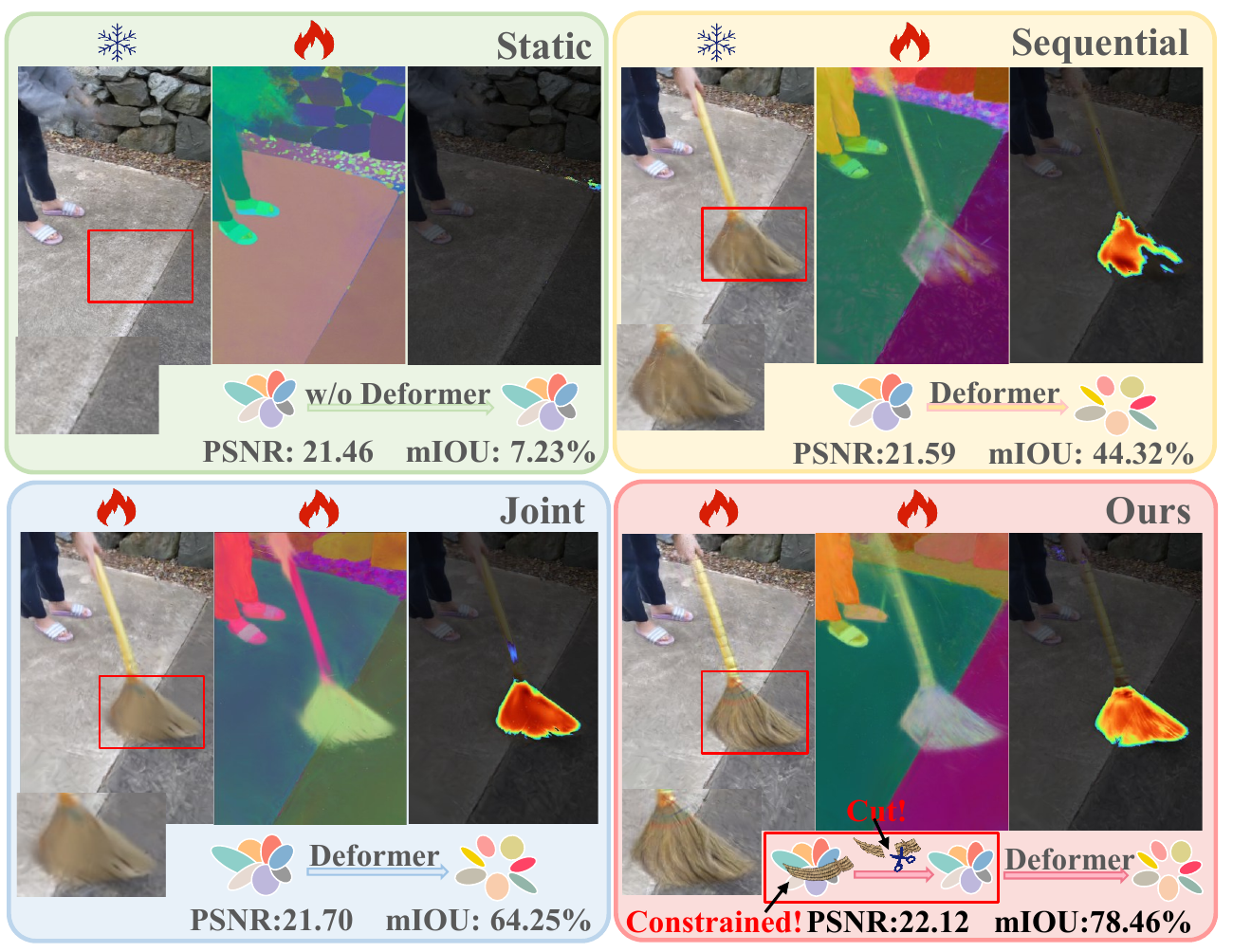}}
\caption{Visualization of different methods. Top Left: Use static methods directly. Top Right: First train the 4D scene and then freeze the geometric and color parameters, adding the semantic feature property for optimization. Bottom Left: Jointly optimizing the 4D scene and semantic features. Bottom Right: Our Dual-Hierarchical Optimization method. By adopting a divide-and-conquer strategy for the Gaussians optimization, we achieve substantial improvements in both rendering quality and segmentation accuracy.}
\label{head}
\end{figure}

An intuitive idea for extending static understanding methods to dynamic scenes is to replace the static 3DGS with dynamic 4DGS. The dynamic scene is firstly trained, subsequently, the geometric and color attributes of the Gaussians are frozen, while semantic attributes are introduced to enhance the representation. However, as illustrated in the upper right of Fig.~\ref{head}, this approach can lead to the incomplete extraction of features. Another idea is to jointly optimize the geometric and semantic properties, however, as shown in the lower left corner of Fig.~\ref{head}, Gaussians tend to blur object texture details to improve semantic understanding completeness, which is also undesirable.
Thus, only one question remains: \textbf{how can we effectively optimize Gaussians representing static backgrounds and dynamic foregrounds to achieve both high-fidelity rendering and accurate semantics?}
We propose the DHO method,  which adopts a divide-and-conquer strategy to simultaneously optimize newly introduced semantic features and geometric attributes of the Gaussians in different stages.

In detail, Our proposed Dual-Hierarchy mainly consists of hierarchical Gaussian flow (HGF) and hierarchical Gaussian guidance (HGG). The former primarily involves hierarchical Gaussian management, assigning different geometric and semantic degrees of freedom to Gaussians at different stages. This enables the effective separation of dynamic objects from the static background and their semantic features. As shown in Fig.~\ref{large_deformation}, we extract the top-k Gaussian points with the largest deformation variables for rendering. The original 4DGS renders images that include a significant amount of static background. Ideally, the Gaussians with large deformation should focus on the dynamic foreground. Our HGF facilitates the separation of dynamic and static Gaussians, enabling dynamic Gaussians to concentrate on the dynamic parts that genuinely require deformation. Our HGG primarily applies layered rendering guidance to different parts of the dynamic scene. By using the prior information of dynamic objects obtained through semantic understanding, we apply greater rendering guidance to the dynamic foreground while appropriately reducing the attention of Gaussians to the background noise. This improves the rendering quality of high-texture dynamic objects in complex backgrounds. Our contributions are as follows:

\begin{itemize}
\item We propose an effective dynamic scene optimization method that can be integrated with any model to produce higher-quality rendering and more accurate semantics.
\item The main component of our approach is Dual-Hierarchy Optimization, which consists of two primary modules: hierarchical Gaussian flow and hierarchical Gaussian guidance. The former enables multi-level geometric and semantic optimization, effectively separating dynamic and static parts. The latter frees a large number of Gaussians from constructing background noise, enhancing reconstruction and comprehension of dynamic foreground.
\item Our method provides a foundation for semantic downstream tasks in 4D scenes, such as semantic segmentation and editing. We demonstrate the superiority of our proposed DHO method via comprehensive experiments.

\end{itemize}

\section{Related Works}
\subsection{3D Rendering}

\begin{figure}[t]
\centerline{\includegraphics[width=0.95\linewidth]
{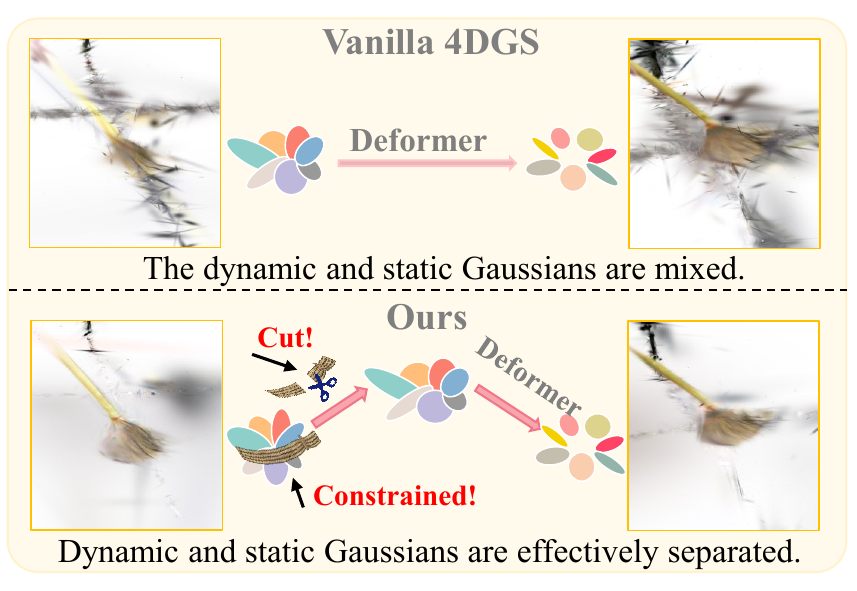}}
\caption{Visualization of Gaussian points with large deformation. We select the top-k Gaussian points with the largest deformation for rendering. Vanilla 4DGS mixes highly deformable dynamic foregrounds with static backgrounds. Our method effectively separates the static and dynamic parts.}
\label{large_deformation}
\end{figure}

The recently developed 3D Gaussian Splatting(3DGS) \cite{3dgs} sets new standards in both rendering quality and speed by using fast differentiable rasterization instead of volume rendering. Significant achievements have been made in dynamic scene reconstruction based on 3DGS, mainly leveraging methods like incremental translation \cite{luiten2023dynamic}, temporal extension \cite{yang2024deformable}, and global deformation \cite{huang2024sc}.
In this paper, we adopt a global deformation Gaussian Splatting representation called 4DGS \cite{wu20244d} for dynamic modeling, which builds on 3DGS by introducing a deformation field to represent the movement and deformation of Gaussians. However, 4DGS cannot focus Gaussians with significant deformations on the dynamic foreground, leading to a mixture of dynamic and static Gaussians, as shown in Fig.~\ref{large_deformation} Our hierarchical Gaussian management effectively facilitates the separation of foreground and background rendering.

\subsection{3D Static Feature Fields}
Progress has also been made in 3D static understanding work based on 3DGS. Gaussian Grouping \cite{gaussian_grouping} uses video tracking methods to bind identical instance objects across different views with the same identity encoding. LangSplat \cite{qin2024langsplat} is the first method to embed CLIP into 3DGS scenes, using a scene language autoencoder to learn language features in the latent space of specific scenes, with rendered semantic features supervised by reduced-dimensional CLIP features. However, these methods are only effective for static scenes, simply extending static methods to dynamic scenes may impact rendering quality and semantic accuracy as shown in Fig.~\ref{head}. 

\subsection{3D Dynamic Feature Fields}

Limited research focuses on dynamic scene understanding. To the best of our knowledge, with two concurrent works known to date. The work most similar to ours is DGD \cite{labe2025dgd}. While demonstrating promising capabilities in simple scenes, it struggles to achieve high-fidelity rendering and accurate semantic features in texture-rich and complex scenes, as shown in Fig.~\ref{result} (a) and (b).
Another approach, SA4D \cite{ji2024segment} has achieved impressive results in 4D scene editing. However, its performance is susceptible to prior video tracking accuracy. As shown in Fig.~\ref{result} (c), when objects in the scene undergo significant deformation or frequent entry and exit, 4D segmentation easily loses small objects. 

\section{Method}

\begin{figure*}[t]
\centerline{\includegraphics[width=0.98\linewidth]
{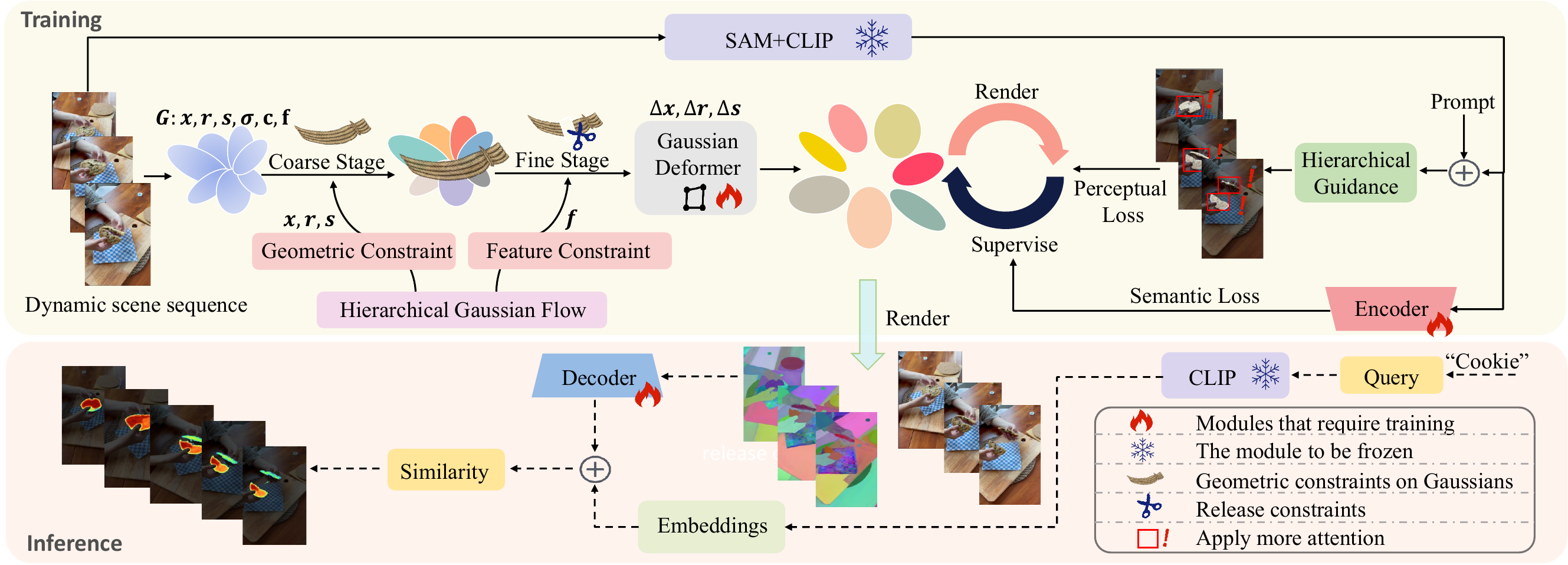}}
\caption{The overall pipeline of our model. We add semantic properties to each Gaussian and obtain the geometric deformation of the Gaussian at each timestamp $t$ through the deformation field. In the coarse stage, Gaussians are subjected to geometric constraints. While in the fine stage, geometries are relaxed and semantic feature constraints are introduced, ensuring foreground-background separation. We utilize dynamic foreground masks obtained from scene priors for hierarchical Gaussian guidance of the scene, enhancing the rendering quality of dynamic foregrounds with complex backgrounds.}
\label{pipeline}
\end{figure*}

\subsection{4D Semantic Gaussian Representation}
We denote the video sequence of a dynamic scene captured by a monocular camera and calibrated by SfM \cite{schonberger2016structure} as $V=\{I_t | t = 1, 2, \ldots, T, I_t \in \mathbb{R}^{3 \times H \times W}\}$, where H and W represent the height and width of the image size. And the initial Gaussians are obtained from the sparse point cloud generated by SfM.
Preserving all properties from the original Gaussian model setup, we introduce additional semantic feature property $\mathbf{f}$, leading to the following representation of the Gaussian $G = \{ \mathbf{\mathcal{X}}_i, \mathbf{s}_i, \mathbf{r}_i, \mathbf{f}_i, \alpha_i, c_i\}$, where the semantic feature $\mathbf{f} \in \mathbb{R}^{n}$.

The semantics of the scene are obtained by segmenting multi-scale objects using SAM \cite{kirillov2023segment}, with semantic information extracted through CLIP \cite{radford2021learning}. Due to the high-dimensional nature of CLIP’s features $\Phi_t^l(p)$, memory issues may arise, and the Gaussians struggle with such high-dimensional features. To address this, we introduce an autoencoder to reduce the dimensionality of the semantics, enabling the Gaussian to work with more compact representations $L_t^l(p)$. The semantic goal is to minimize the difference between $L_t^l(p)$ and $F_t^l(p)$.\begin{equation}
\mathcal{L}_s = \sum_{l \in \{s, m, l\}} \sum_{t=1}^{T} \mathcal{L} \left( L_t^l(p), F_t^l(p) \right), \label{eq4}
\end{equation}
where $s$, $m$, $l$ represent different scales of the feature, from subpart, part to the whole. In the semantic query process, a decoder  $D$ raises the low-dimensional Gaussian semantic feature $F_t^l(p)$ to the original CLIP dimension. And encoder $E$ and decoder $D$ should be ensured to minimize the loss of feature information in the process of changing dimensions.
\begin{equation}
\mathcal{L}_{\text{auto}} = \sum_{l \in \{s, m, l\}} \sum_{t=1}^{T} \mathcal{L} \left( D \left(E \left(\Phi_t^l(p) \right) \right), \Phi_t^l(p) \right). \label{eq5}
\end{equation}

The deformation field $\mathcal{F}$ calculates the deformation offsets of the Gaussian points. We can get the Gaussian set $G_t = \{ \mathbf{\mathcal{X}}_i + \Delta \mathbf{\mathcal{X}}, \mathbf{r}_i+ \Delta \mathbf{r}, \mathbf{s}_i+ \Delta \mathbf{s}, \mathbf{f}_i, \alpha_i, c_i\}$ at the timestamp $t$, the additional semantic features $\mathbf{f}$ are bound to each Gaussian and are not deformed by the deformation field. Upon projecting the 3D Gaussians into a 2D space, the color $C$ of a pixel and the semantic feature value $F$ of a feature map pixel are computed by volumetric rendering which is performed using front-to-back depth order:
\begin{equation}
C(v) = \sum_{i \in \mathcal{N}} c_i \alpha_i \prod_{j=1}^{i-1} (1 - \alpha_j),
\end{equation}

\begin{equation}
F(v) = \sum_{i \in \mathcal{N}} f_i \alpha_i \prod_{j=1}^{i-1} (1 - \alpha_j),
\end{equation}
where $C(v)$ is the rendered color at pixel $v$, $F(v)$ represents the semantic feature embedding rendered at pixel $v$. 

\subsection{Dual-Hierarchical Optimization }

\noindent\textbf{Hierarchical Gaussian Flow.} As an explicit representation similar to point clouds, Gaussian representations suffer from instability in each iterative update, especially in scenes with both dynamic and static components. Inspired by GaussianEditor \cite{chen2024gaussianeditor}, we propose the Hierarchical Gaussian Flow (HGF), which applies different degrees of freedom to the geometric and semantic properties of Gaussians at various stages. In the coarse stage, we aim to stabilize the geometry of Gaussians for static background, imparting lower degrees of freedom to them. In the fine stage, Gaussians primarily reconstruct dynamic objects in the scene. Due to the significant deformations and richer texture details of dynamic objects, we allow more freedom in these Gaussians to adequately capture the dynamic details. However, we impose certain constraints on semantic features to enhance their cohesion, ensuring effective separation from static background features. This approach contributes to the integrity and compactness of semantic features.

During the training process, points generated in the $i$-th densification round are marked as generation $i$. The older the generation, the stronger the constraints applied. We apply anchor loss to impose different geometric and semantic constraints on Gaussians across different generations to control their flexibility. Earlier-introduced Gaussians in the background receive stronger geometric constraints in the coarse stage, while those in the foreground receive stronger feature constraints in the fine stage. At the beginning of training, HGF records the attributes of all Gaussians as anchors, which are updated to reflect the current state of each Gaussian during each densification process. Throughout the training, Mean Squared Error (MSE) loss between anchor states and current states is employed to ensure that Gaussians do not deviate too far from their respective anchors.

\begin{equation}
\mathcal{L}^s_{anchor} = \sum_{i=0}^n \lambda_i [(S_{i, c} * \theta_{c} + S_{i, f} * \theta_{f}) - \hat{S}_i]^2,
\end{equation}
where $S_{i, c}$ denotes a certain property of the Gaussian in the coarse stage ($\theta_{c}=1,\theta_{f}=0$) from $\{\mathbf{\mathcal{X}}_i, \mathbf{s}_i, \mathbf{r}_i, \alpha_i, c_i\}$. And $S_{i, f}$ represents $\mathbf{f}_i$ in the fine stage ($\theta_{c}=0,\theta_{f}=1$). $\hat{S}$ refers to the same property recorded in the anchor state. $\lambda_i$ indicates the strength of the anchor loss applied to the $i$-th Gaussian, which is determined by its generation process.

\noindent\textbf{Hierarchical Gaussian Guidance.} In dynamic scenes with noisy backgrounds, Gaussians can easily mistake noise for important details, which lowers the rendering quality of dynamic foreground objects. For instance, the ``Broom" scenario in the HyperNeRF dataset, numerous black dots on the ground consume much of the Gaussians resources, leading to distortion and blurriness in rendering the dynamic broom parts. Obviously, we are more concerned with the rendering of the foreground objects in these scenes. The original 4DGS, lacking semantic features as guidance, faces challenges in distinguishing areas that require more focus.

Semantic prompts are provided for dynamic objects that need refinement in the scene, then utilize CLIP for foreground mask  $M$. And we decompose the scene image $I_t$ into two parts: those within the mask and those outside it:
\begin{equation}
I_m = I_t \cdot M, \quad I_{\overline{m}} = I_t \cdot (1 - M).
\end{equation}
From the refined mask, our 4D semantic Gaussians learn the hierarchical Gaussian guidance.
\begin{equation}
\mathcal{L}_{\text{mask}}^v = \lambda_m \mathcal{L}(I_m - \hat{I}_m) + (1-\lambda_m)\mathcal{L}(I_{\overline{m}} - \hat{I}_{\overline{m}}), \label{eqL_mask}
\end{equation}

where $\hat{I}$ is the rendered image, $\lambda_m$ is the adaptive weight smoothed and constrained using the Sigmoid function $\lambda_m=\sigma\left(\alpha (\bar{D} - \beta)\right)$, adjusted based on the scene's average texture density $\bar{D}$ to control the attention of Gaussians inside and outside the mask. We use the mean texture density $\bar{D}$ \cite{niimi1997image} to assess the richness of scene texture, which is the average of the proportion of edge pixels to the total pixels in the scene:
\begin{equation}
\bar{D} = \frac{1}{N} \sum_{k=1}^{N} \frac{\sum_{i=1}^{H_k} \sum_{j=1}^{W_k} E_{ij}^{(k)}}{H_k \times W_k},
\end{equation}
where $k$ represents the $k$-th image, $N$ is the total number of scene images captured, $H_k$ and $W_k$ are the height and width of the $k$-th image respectively. $E_{ij}^{(k)}$ denotes the edge detection result at position $(i, j)$ in the $k$-th image.

\noindent\textbf{Training Strategy.} 
Firstly, we train an autoencoder with \eqref{eq5} and then use \eqref{eq4} as a semantic loss function $\mathcal{L}_s$. A grid-based total-variational loss $\mathcal{L}_{\text{tv}}$ \cite{cao2023hexplane} is also applied. The total loss is:
\begin{equation}
\mathcal{L} = \lambda_m\mathcal{L}_{\text{mask}}^v + \lambda_a\mathcal{L}_{\text{anchor}}^s + \lambda_s\mathcal{L}_s + \lambda_{\text{tv}}\mathcal{L}_{\text{tv}}. \label{eq13}
\end{equation}

\section{Experiments}
\subsection{Experimental Setups}
\noindent\textbf{Implementation Details.} Our implementation is primarily based on the PyTorch and tested in a single RTX 3090 GPU. During the coarse stage, the static background is pre-trained for 5k iterations without hierarchical Gaussian guidance. Deformation field training is introduced in the fine stage. We use the Adam optimizer with 30K training iterations, which takes approximately one hour. 

\noindent\textbf{Datasets.} We use a synthetic dataset of D-NeRF \cite{pumarola2021d} and real-world datasets of HyperNeRF \cite{park2021hypernerf} and Neu3D’s \cite{li2022neural} to evaluate our method. Since the datasets lack segmentation ground truth, they are unsuitable for direct testing. And the parallel DGD work \cite{labe2025dgd} does not expose its annotation results and semantic segmentation code. Therefore, we consider performing manual annotation, we begin by applying SAM to each frame and then manually refine the provided segments, as illustrated in the supplementary materials.
\begin{figure*}[t]
\centerline{\includegraphics[width=0.98\linewidth]
{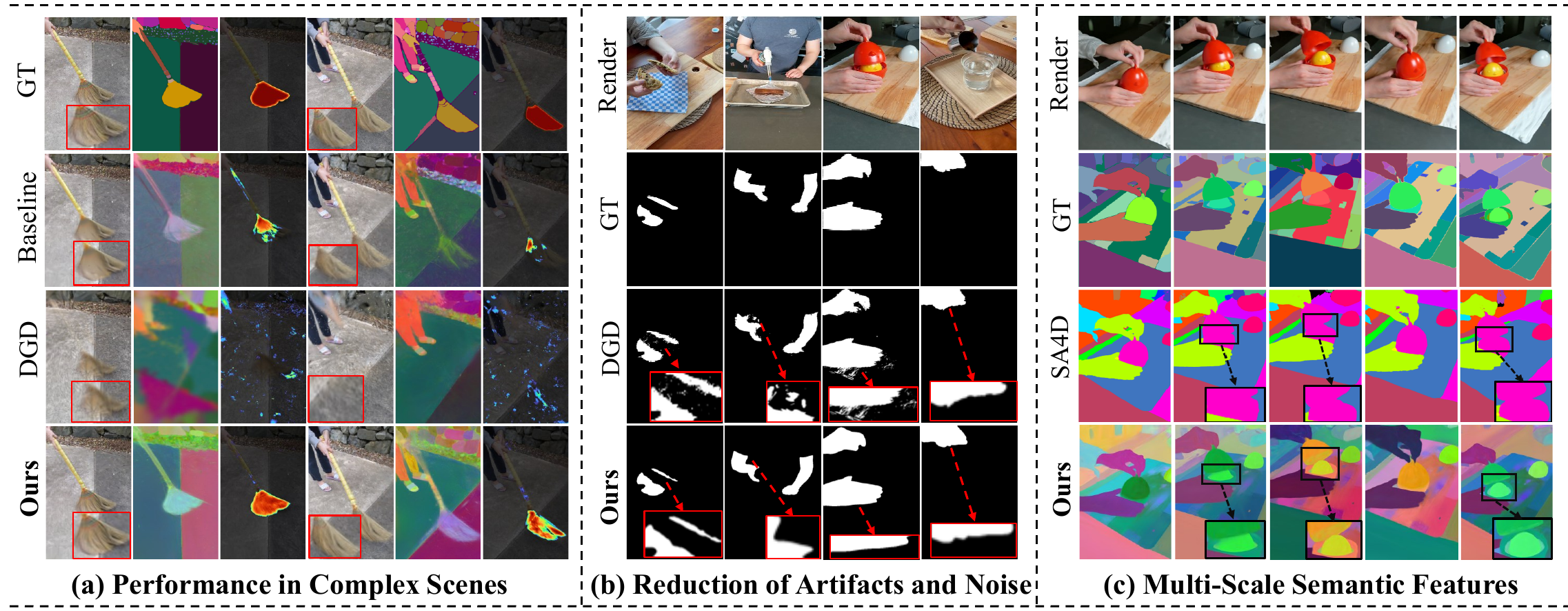}}
\caption{Visualizatio of the HyperNeRF dataset. (a) Visualization of the ``Broom" scene. Our method outperforms the baseline, whereas DGD nearly fails in complex scenes. (b) Visual segmentation comparisons of our method and DGD. Our method significantly reduces artifacts and noise. (c) Visual scale comparisons of our method and SA4D. Our method incorporates multi-scale information to perceive objects at different scales.}
\label{result}
\end{figure*}

\noindent\textbf{Baselines.} Due to the recent lack of work on 4D semantic understanding, apart from DGD and SA4D, we also consider two additional categories of baselines. (1) One approach is sequential training, where we first pre-train the 4DGS to reconstruct the dynamic scene, then freeze the Gaussian geometry properties and add semantic feature attributes for further training. (2) The other approach is joint training, which simultaneously optimizes scene geometry and semantic features, without any additional strategic improvements. 

\subsection{Experimental Results}
\noindent\textbf{Rendering and Semantic Results.} Tab.~\ref{tab:render} provides the rendering quality metrics on the HyperNeRF \cite{park2021hypernerf} and D-NeRF \cite{pumarola2021d} datasets, evaluated using PSNR, SSIM, and LPIPS. Due to the small pixel proportion of the dynamic foreground in dynamic scenes, the overall scene metrics show limited improvement. To further highlight the quality enhancement in the dynamic foreground, we specifically compute metrics for the local dynamic region, as shown in parentheses in Tab.~\ref{tab:render}, with global metrics provided outside parentheses. Tab.~\ref{tab:segment} shows our mIoU metrics. The results show that our method outperforms the baseline in both rendering quality and semantic segmentation accuracy, particularly in improving the dynamic foreground quality in complex scenes. Our method can be integrated with any model, achieving significant performance improvements, as shown in Tab.~\ref{tab:models}. 

Specifically, as shown in Fig.~\ref{result} (a), for the finely textured scene, the baseline performs poorly in rendering the broom's foreground. By employing the DHO, our method results in more detailed textures of the broom hair and accurately depicts the colorful stripes on the handle. 
Particularly, our method shows significant improvements in terms of segmentation accuracy. Both the baseline and DGD methods perform poorly, while our method extracts complete semantic features, effectively separating dynamic object semantics from the static background, enabling precise localization of specified objects, with more concentrated heatmaps.
As shown in Fig.~\ref{result} (b), DGD produces many artifacts between static background and dynamic foreground. Our method significantly reduces the noise. As shown in Fig.~\ref{result} (c), our method incorporates multi-scale information to perceive multi-scale objects. However, SA4D \cite{ji2024segment} fails to capture small deformable objects. 

\begin{table}[t]
\caption{Quantitative results on the D-NeRF (left) and HyperNeRF (right) datasets. Metrics are reported globally (outside parentheses) and on the local dynamic foreground (inside parentheses). Our method outperforms the baseline on both real-world and synthetic datasets.}
    \begin{center}
\renewcommand\arraystretch{0.5}
    \begin{tabular}{lcccc}
        \toprule
        \textbf{Model}
        & \textbf{PSNR(dB)↑} & \textbf{SSIM↑} & \textbf{LPIPS↓} \\
        \midrule
         Sequential & 34.50\textbar28.09(28.44) & 0.97\textbar0.78(0.95) & 0.018\textbar0.29(0.038) \\
         Joint & 34.84\textbar28.12(28.82) & 0.97\textbar0.79(0.96) & 0.013\textbar0.28(0.031) \\
         Ours & \textbf{35.29}\textbar\textbf{28.41(29.86)} & \textbf{0.98}\textbar\textbf{0.82(0.98)} & \textbf{0.009}\textbar\textbf{0.24(0.023)} \\
        \bottomrule
    \end{tabular}
    \label{tab:render}
    \end{center}
\end{table}

\begin{table}[h]
\centering
\caption{Quantitative results of combining different dynamic reconstruction methods with ours on the Neu3D’s dataset. Outside the parentheses: original model, inside: model with our strategy. Our method demonstrates the ability to be integrated into arbitrary models, leading to enhanced rendering quality and improved semantic accuracy.}
\label{tab:models}
    \resizebox{0.48\textwidth}{!}{
    \large
    \begin{tabular}{*{10}{c}}
        \toprule
       \textbf{Model} & \textbf{PSNR(dB)↑} &  \textbf{D-SSIM↓} & \textbf{LPIPS↓} &  \textbf{mIoU(\%)↑}  \\ 
        \midrule
        4DGS + (Ours) & 31.02 (\textbf{31.96}) & 0.015 (\textbf{0.013}) & 0.049 (\textbf{0.034}) &  78.14 (\textbf{88.35})  \\ 
       D-3DGS \cite{yang2024deformable}  + (Ours) & 30.65 (\textbf{31.52}) & 0.018 (\textbf{0.016}) & 0.056 (\textbf{0.042}) &  73.25 (\textbf{89.23}) \\
        SC-GS \cite{huang2024sc} + (Ours) & 30.24 (\textbf{31.33}) & 0.017 (\textbf{0.016}) & 0.058 (\textbf{0.045})&  81.33 (\textbf{90.07})   \\
        \bottomrule
    \end{tabular}
    }

\end{table}

\begin{table}[t]
\caption{mIoU (\%) metric for object segmentation on the D-NeRF and HyperNeRF datasets. Due to the unavailability of DGD's annotation and segmentation code, we can only use the results presented in their paper. Our method achieves higher segmentation accuracy and significantly reduces artifacts and noise, as shown in Fig.~\ref{result} (b).}
    \begin{center}
    \setlength{\tabcolsep}{7pt}
    \renewcommand\arraystretch{0.5}
    \begin{tabular}{lcccc}
        \toprule
          &\textbf{Sequential} & \textbf{Joint} & \textbf{DGD$^*$} & \textbf{Ours} \\ 
        \midrule
        D-NeRF & 63.76 & 83.80 & 77.51 & \textbf{91.04}\\
        HyperNeRF & 70.54 & 78.14 & 77.88 & \textbf{88.35}\\
        \bottomrule
    \end{tabular}
    \label{tab:segment}
    \end{center}
\end{table}

\begin{figure}[t]
\centerline{\includegraphics[width=0.95\linewidth]
{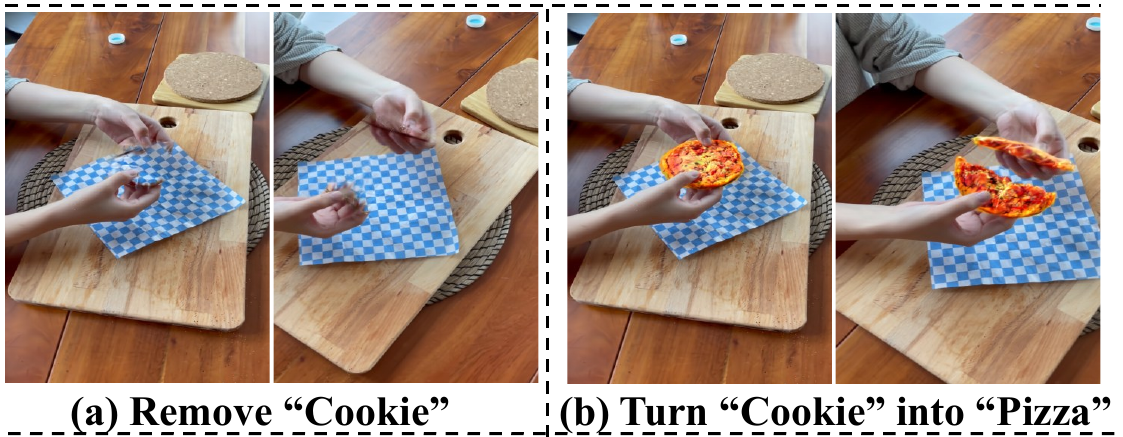}}
\caption{Visualization of object editing. Our language prompts provide a more intuitive way of object selection than ambiguous IDs of SA4D.}
\label{remove}
\end{figure}

\noindent\textbf{Semantic Editing.} Our method also supports other downstream tasks. As shown in Fig.~\ref{remove}, it enables dynamic object removal and editing. Detailed implementation and results are provided in the supplementary materials. By providing semantic prompts, our approach offers a more intuitive way of object selection than SA4D, which requires a unique ID, and the correspondence between IDs and objects is not predetermined.

\subsection{Ablation Studies}
\begin{figure}[t]
\centerline{\includegraphics[width=0.98\linewidth]
{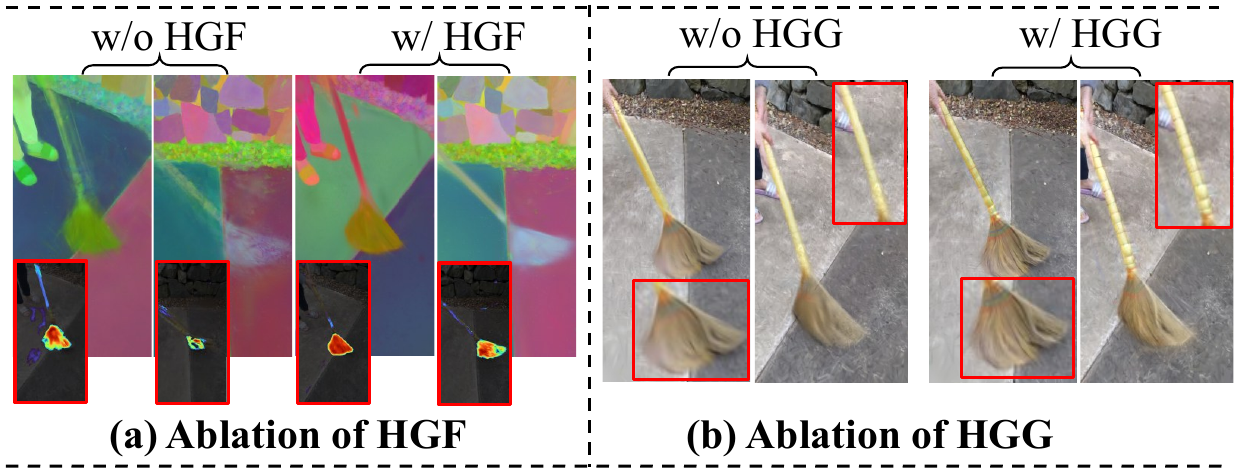}}
\caption{Visualization of ablation study. (a) Ablation of HGF results in fragmented semantic features. (b) Ablation of HGG results in degradation in foreground rendering quality.}
\label{ab}
\end{figure}

\noindent\textbf{Hierarchical Gaussian Flow.} The HGF plays a pivotal role in the separation of dynamic foreground Gaussians and static background Gaussians. As shown in Tab.~\ref{tab:ablation}, the ablation of HGF greatly affects the segmentation accuracy. As illustrated in Fig.~\ref{ab} (a), eliminating HGF from the model architecture results in fragmented semantic features, adversely impacting the integrity of the confidence heatmaps for segmentation.

\begin{table}[!t]
\caption{Quantitative results of ablation study. Metrics are reported globally (outside parentheses) and on the local dynamic foreground (inside parentheses), our method improves overall performance while significantly enhancing results within dynamic regions. (a) Ablation of HGF significantly affects the semantic segmentation accuracy. (b) Ablation of HGG results in degraded rendering quality of the foreground.}
    \centering
    \setlength{\tabcolsep}{2pt}
    \renewcommand\arraystretch{0.5}
    \begin{tabular}{lcccc}
        \toprule
        \textbf{Model}
        & \textbf{PSNR(dB)↑} & \textbf{SSIM↑} & \textbf{LPIPS↓} & \textbf{mIoU↑} \\ 
        \midrule
        w/o HGF & 21.83 (29.21) & 0.35 (0.96) & 0.54 (0.027) & 70.32\\
        w/o HGG & 21.79 (29.09) & 0.34 (0.97) & 0.55 (0.031) & 72.46\\
        w/o HGF\&HGG & 21.70 (28.44) & 0.34 (0.95) & 0.56 (0.038) & 64.25\\
        Ours & \textbf{22.12} (\textbf{29.86}) & \textbf{0.39} (\textbf{0.98}) & \textbf{0.47} (\textbf{0.023}) & \textbf{78.46}\\   
        \bottomrule
    \end{tabular}
    \label{tab:ablation}
\end{table}

\begin{table}[t]
\caption{Quantitative results of adaptive adjustment of $\lambda_m$ across different scenes (Left: Without HGG, Right: With adaptive adjustment). Our method adaptively adjusts the value of $\lambda_m$.}
    \centering
    \setlength{\tabcolsep}{7pt}   \renewcommand\arraystretch{0.3}
    \begin{tabular}{lcccc}
        \toprule
        \textbf{$\bar{D}$} & \textbf{$\lambda_m$} & \textbf{PSNR(dB)↑}  & \textbf{mIoU↑} \\
        \midrule
        0.21 (Broom Scene) & 0.85 & 21.70\textbar\textbf{22.67} & 64.25\textbar\textbf{78.46} \\
        0.05 (Cookie Scene) & 0.60 & 32.70\textbar\textbf{33.55} & 74.55\textbar\textbf{92.19} \\
        0.02 (Hook Scene)& 0.55 & 34.50\textbar\textbf{35.31}  & 76.01\textbar\textbf{87.49} \\
        
        \bottomrule
    \end{tabular}
    
    \label{tab:lambda_values}
\end{table}

\noindent\textbf{Hierarchical Gaussian Guidance.} The HGG demonstrates remarkable efficacy in enhancing the rendering of dynamic foregrounds in complex backgrounds by leveraging the scene prior. As shown in Tab.~\ref{tab:ablation}, melting HGG results in degraded rendering quality. As depicted in Fig.~\ref{ab} (b), ablating HGG, which implies treating all parts of the scene equally, leads to poor reconstruction performance in noisy backgrounds, with fine texture details of the dynamic foreground becoming blurred. Furthermore, our method enables adaptive adjustment based on the scene's textural complexity, as shown in Tab.~\ref{tab:lambda_values}.

\section{Conclusion}
To tackle the intertwined challenges of dynamic scene reconstruction and understanding, we propose a novel method DHO, which enriches 4DGS with semantics. It employs a divide-and-conquer strategy to optimize Gaussians representing dynamic and static scene parts independently, enhancing the representation of dynamic scenes. Our method efficiently generates novel scene views and precise semantic features, offering robust support for downstream tasks like semantic segmentation and object editing.
\section{Acknowledgments}
This work was supported by grants from the National Natural Science Foundation of China (62301228, 62176100,  62376011). The computation is completed in the HPC Plat
form of Huazhong University of Science and Technology.

\bibliographystyle{IEEEbib}
\bibliography{icme2025references}

\begin{thebibliography}{10}

\bibitem{zhou2024drivinggaussian}
Xiaoyu Zhou, Zhiwei Lin, Xiaojun Shan, Yongtao Wang, Deqing Sun, and Ming-Hsuan Yang,
\newblock ``Drivinggaussian: Composite gaussian splatting for surrounding dynamic autonomous driving scenes,''
\newblock in {\em CVPR}, 2024.

\bibitem{xie2024physgaussian}
Tianyi Xie, Zeshun Zong, Yuxing Qiu, Xuan Li, Yutao Feng, Yin Yang, and Chenfanfu Jiang,
\newblock ``Physgaussian: Physics-integrated 3d gaussians for generative dynamics,''
\newblock in {\em CVPR}, 2024.

\bibitem{3dgs}
Bernhard Kerbl, Georgios Kopanas, Thomas Leimk{\"u}hler, and George Drettakis,
\newblock ``3d gaussian splatting for real-time radiance field rendering,''
\newblock {\em ACM Transactions on Graphics}, 2023.

\bibitem{yi2024gaussiandreamer}
Taoran Yi, Jiemin Fang, Junjie Wang, Guanjun Wu, Lingxi Xie, Xiaopeng Zhang, Wenyu Liu, Qi~Tian, and Xinggang Wang,
\newblock ``Gaussiandreamer: Fast generation from text to 3d gaussians by bridging 2d and 3d diffusion models,''
\newblock in {\em CVPR}, 2024.

\bibitem{qin2024langsplat}
Minghan Qin, Wanhua Li, Jiawei Zhou, Haoqian Wang, and Hanspeter Pfister,
\newblock ``Langsplat: 3d language gaussian splatting,''
\newblock in {\em CVPR}, 2024.

\bibitem{wu20244d}
Guanjun Wu, Taoran Yi, Jiemin Fang, Lingxi Xie, Xiaopeng Zhang, Wei Wei, Wenyu Liu, Qi~Tian, and Xinggang Wang,
\newblock ``4d gaussian splatting for real-time dynamic scene rendering,''
\newblock in {\em CVPR}, 2024.

\bibitem{luiten2023dynamic}
Jonathon Luiten, Georgios Kopanas, Bastian Leibe, and Deva Ramanan,
\newblock ``Dynamic 3d gaussians: Tracking by persistent dynamic view synthesis,''
\newblock in {\em 3DV}, 2024.

\bibitem{yang2024deformable}
Ziyi Yang, Xinyu Gao, Wen Zhou, Shaohui Jiao, Yuqing Zhang, and Xiaogang Jin,
\newblock ``Deformable 3d gaussians for high-fidelity monocular dynamic scene reconstruction,''
\newblock in {\em CVPR}, 2024.

\bibitem{huang2024sc}
Yi-Hua Huang, Yang-Tian Sun, Ziyi Yang, Xiaoyang Lyu, Yan-Pei Cao, and Xiaojuan Qi,
\newblock ``Sc-gs: Sparse-controlled gaussian splatting for editable dynamic scenes,''
\newblock in {\em CVPR}, 2024.

\bibitem{gaussian_grouping}
Mingqiao Ye, Martin Danelljan, Fisher Yu, and Lei Ke,
\newblock ``Gaussian grouping: Segment and edit anything in 3d scenes,''
\newblock in {\em ECCV}, 2024.

\bibitem{labe2025dgd}
Isaac Labe, Noam Issachar, Itai Lang, and Sagie Benaim,
\newblock ``Dgd: Dynamic 3d gaussians distillation,''
\newblock in {\em ECCV}, 2025.

\bibitem{ji2024segment}
Shengxiang Ji, Guanjun Wu, Jiemin Fang, Jiazhong Cen, Taoran Yi, Wenyu Liu, Qi~Tian, and Xinggang Wang,
\newblock ``Segment any 4d gaussians,'' 2024.

\bibitem{schonberger2016structure}
Johannes~L Schonberger and Jan-Michael Frahm,
\newblock ``Structure-from-motion revisited,''
\newblock in {\em CVPR}, 2016.

\bibitem{kirillov2023segment}
Alexander Kirillov, Eric Mintun, Nikhila Ravi, Hanzi Mao, Chloe Rolland, Laura Gustafson, Tete Xiao, Spencer Whitehead, Alexander~C Berg, Wan-Yen Lo, et~al.,
\newblock ``Segment anything,''
\newblock in {\em ICCV}, 2023.

\bibitem{radford2021learning}
Alec Radford, Jong~Wook Kim, Chris Hallacy, Aditya Ramesh, Gabriel Goh, Sandhini Agarwal, Girish Sastry, Amanda Askell, Pamela Mishkin, Jack Clark, et~al.,
\newblock ``Learning transferable visual models from natural language supervision,''
\newblock 2021.

\bibitem{chen2024gaussianeditor}
Yiwen Chen, Zilong Chen, Chi Zhang, Feng Wang, Xiaofeng Yang, Yikai Wang, Zhongang Cai, Lei Yang, Huaping Liu, and Guosheng Lin,
\newblock ``Gaussianeditor: Swift and controllable 3d editing with gaussian splatting,''
\newblock in {\em CVPR}, 2024.

\bibitem{niimi1997image}
Michiharu Niimi, Hideki Noda, and Eiji Kawaguch,
\newblock ``An image embedding by a complexity based region segmentation,''
\newblock in {\em ICIP}, 1997.

\bibitem{cao2023hexplane}
Ang Cao and Justin Johnson,
\newblock ``Hexplane: A fast representation for dynamic scenes,''
\newblock in {\em CVPR}, 2023.

\bibitem{pumarola2021d}
Albert Pumarola, Enric Corona, Gerard Pons-Moll, and Francesc Moreno-Noguer,
\newblock ``D-nerf: Neural radiance fields for dynamic scenes,''
\newblock in {\em CVPR}, 2021.

\bibitem{park2021hypernerf}
Keunhong Park, Utkarsh Sinha, Peter Hedman, Jonathan~T. Barron, Sofien Bouaziz, Dan~B Goldman, Ricardo Martin-Brualla, and Steven~M. Seitz,
\newblock ``Hypernerf: A higher-dimensional representation for topologically varying neural radiance fields,''
\newblock {\em ACM Transactions on Graphics}, 2021.

\bibitem{li2022neural}
Tianye Li, Mira Slavcheva, Michael Zollhoefer, Simon Green, Christoph Lassner, Changil Kim, Tanner Schmidt, Steven Lovegrove, Michael Goesele, Richard Newcombe, et~al.,
\newblock ``Neural 3d video synthesis from multi-view video,''
\newblock in {\em CVPR}, 2022.

\end{thebibliography}

\newpage
\twocolumn[
\begin{@twocolumnfalse}
\begin{center}
\textbf{\LARGE Supplementary Material}
\vspace{2em}
\end{center}
\end{@twocolumnfalse}
]

\setcounter{section}{0}

\section*{Summary}
\label{sec:intro}
This supplementary material is organized as follows.
\begin{itemize}
\item Section~\ref{supp_sec2} provides more details of the experiments.
\item Section~\ref{supp_sec3} provides more quantitative results.
\item Section~\ref{supp_sec4} describes the comparison results between DGD method and ours.
\item Section~\ref{supp_sec5} shows the multi-scale semantic features of ours.
\item Section~\ref{supp_sec6} provides more visualization results.
\item Section~\ref{supp_sec7} introduces the segmentation mask annotation.

\end{itemize}

\begin{table}[!htpb]
\caption{Quantitative results on the relationship between training iterations and performance.}
    \begin{center}
    \renewcommand\arraystretch{0.8}
    \begin{tabular}{lcccc}
        \toprule
        \textbf{Iterations}
        & \textbf{PSNR(dB)↑} & \textbf{mIoU↑} & \textbf{Time↓}
        & \textbf{Storage↓}\\
        \midrule
         3k+10k & 32.01 &82.77\% & \textasciitilde30min & 47\\
         5k+15k & 32.52 & 88.35\% & \textasciitilde1h &58 \\
         5k+20k & 32.58& 89.12\% & \textgreater1h30min&72 \\
        \bottomrule
    \end{tabular}
    \label{tab:time}
    \end{center}
\end{table}

\begin{table}[!t]
\caption{Quantitative results of loss function under various scaling factors on the HyperNeRF dataset. Our method shows strong robustness to different scaling factors, as long as the magnitudes of loss components are comparable.}
    \centering
   \renewcommand\arraystretch{0.8}
    \resizebox{\linewidth}{!}{
    \begin{tabular}{lcccccc}
        \toprule
        & \textbf{$\lambda_m$} & \textbf{$\lambda_a$} & 
        \textbf{$\lambda_s$} &
    \textbf{$\lambda_{\text{tv}}$} &
        \textbf{PSNR(dB)}  & \textbf{mIoU} \\
        \midrule
        &1 & 1 & 0.1 & 1 & 28.41 & 88.35\\
        &0.8 & 1 & 0.1 & 1  & 28.43 & 87.98 \\
        &1 & 0.8 & 0.1 & 1 & 28.39 & 88.11 \\
        \bottomrule
    \end{tabular}}
    \label{tab:loss_function}
\end{table}

\section{More Implementation Details}
\label{supp_sec1}
\noindent\textbf{Settings.} We fine-tune our optimization parameters by the configuration outlined in the 4DGS. Our dual-hierarchical optimization method performs a stage-wise optimization for 5k iterations in the coarse phase and 15k iterations in the fine phase. We have experimented with different numbers of optimization iterations, as shown in Tab.~\ref{tab:time}.  While performance improves with more training iterations, the gains become less significant after 15k iterations.  Additionally, as the number of iterations increases, both training time and storage requirements grow. To balance efficiency and performance, all the results reported in the paper are based on 5k+15k iterations. 

For (9), we set $\lambda_m=1$, $\lambda_a=1$, $\lambda_s=0.1$, $\lambda_{\text{tv}}=1$, because the semantic loss $\lambda_s$ is typically an order of magnitude larger than other loss parts. As shown in Tab.~\ref{tab:loss_function}, experiments with various values of $\lambda_m$, $\lambda_a$, $\lambda_s$ and $\lambda_{\text{tv}}$ demonstrate the robustness of our loss function under different scaling factors. The autoencoder compresses the extracted CLIP features $\Phi_t^l(p) \in \mathbb{R}^{512}$ into $L_t^l(p) \in \mathbb{R}^{8}$, our semantic Gaussians learn the semantic feature $f \in \mathbb{R}^{8}$. Baselines and our method both learn the same semantic feature dimensions and we perform PCA on semantic feature map $F_t^l(p) \in \mathbb{R}^{8}$ to ensure fairness. The learning rate is set as $1.6 \times 10^{-3}$, decayed to $1.6 \times 10^{-4}$ at the end of training. The Gaussian deformation decoder is a tiny MLP with a learning rate of $1.6 \times 10^{-3}$. The training batch size is 2. The opacity reset operation is not used as it does not bring evident benefit in most of our tested scenes.

For $\lambda_m=\sigma\left(\alpha (\bar{D} - \beta)\right)$, we set $\alpha=10$, $\beta=0.01$. $\alpha$ and $\beta$ are adjustable, with 
$\alpha=10$ and $\beta=0.01$ being effective for most scenes through extensive experiments. The value of $\lambda_m$ is adaptively adjusted according to scene complexity $\bar{D}$, influencing the attention of Gaussians to different parts of the dynamic scene. This helps mitigate background noise interference and improves the rendering quality of the dynamic foreground. The number of edge pixels in (8) can be calculated using different edge detection algorithms, we use the canny operator in the paper. We choose Canny, Sobel, LOG, and Laplacian operators for testing. Different edge detection algorithms exhibit similar performance in judging scene complexity. Thus, we can choose any algorithm for edge detection, but it is crucial to ensure consistency in the algorithm used across different scenes. 

\noindent\textbf{Semantic Querying.} Semantic queries are processed using the LERF  to calculate relevance scores for each text query. 
\begin{equation}
T_s=\min_i \frac{\exp(\phi_{emb} \cdot \phi_{qry})}{\exp(\phi_{emb} \cdot \phi_{qry}) + \exp(\phi_{emb} \cdot \phi_{canon}^i)},
\end{equation}
where $\phi_{emb}$ is rendered language embedding, $\phi_{qry}$ is text query, $\phi_{canon}^i$ is a pre-defined canonical stage CLIP embedding selected from a set of predefined terms including \textit{``object"}, \textit{``things"}, \textit{``stuff"}, and \textit{``texture"}. Following the LERF strategy, we select the semantic level yielding the highest relevance score. For 4D semantic segmentation, we filter out points with relevance scores below a chosen threshold and predict object masks for the remaining regions.

\noindent\textbf{Semantic Editing.} We select Gaussian points with similar semantic similarities $T_e$.
\begin{equation}
T_e(L_t^l(p), F_t^l(p)) = \frac{L_t^l(p)\cdot F_t^l(p)}{\|L_t^l(p)\| \|F_t^l(p)\|}.
\label{eqL_mask}
\end{equation}
When the semantic similarity $T_e$ exceeds the predefined threshold $\sigma$, we group the Gaussians that are semantically similar to the prompt $i$ and edit only these Gaussians, while freezing the others. The Gaussians within the group are optimized by freezing their geometric properties and training only their color attributes. 
We follow the DreamFusion and optimize using the SDS as the editing loss, we can efficiently edit dynamic objects in the scene based on their semantics.
\begin{equation}
\nabla_{\theta} \mathcal{L}_{\text{SDS}} (\phi, x) = w(t) \left[ \left( \hat{\epsilon}_{\phi}(x_t; y, t) - \epsilon \right) \right] \frac{\partial x}{\partial \theta}.
\end{equation}

\begin{table}[!t]
\caption{Quantitative results of rendered image quality for various scenes in the HyperNeRF dataset. Sequential Baseline (Left), Joint Baseline (Middle), Ours (Right). Our method outperforms the baselines on real-world dataset.}
    \centering
    \resizebox{\linewidth}{!}{
    \begin{tabular}{lccc}
        \toprule
        \textbf{Scene}
        & \textbf{PSNR(dB)↑} & \textbf{SSIM↑} & \textbf{LPIPS↓} \\
        \midrule
        Broom & 21.59\textbar21.70\textbar\textbf{22.12} & 0.35\textbar0.36\textbar\textbf{0.39} & 0.56\textbar0.53\textbar\textbf{0.47} \\
        Cookie & 32.77\textbar32.83\textbar\textbf{33.55} & 0.90\textbar0.92\textbar\textbf{0.93} & 0.09\textbar0.08\textbar\textbf{0.06} \\
        Lemon & 30.31\textbar30.45\textbar\textbf{30.76} & 0.72\textbar0.78\textbar\textbf{0.81} & 0.26\textbar0.22\textbar\textbf{0.16} \\
        Chicken & 26.96\textbar26.89\textbar\textbf{27.16} & 0.78\textbar0.77\textbar\textbf{0.79} & 0.20\textbar0.18\textbar\textbf{0.16} \\
        Chocolate & 27.61\textbar27.65\textbar\textbf{27.98} & 0.86\textbar0.88\textbar\textbf{0.90} & 0.96\textbar0.96\textbar\textbf{0.94} \\
        Americano & 31.38\textbar31.16\textbar\textbf{31.47} & 0.91\textbar0.91\textbar\textbf{0.92} & 0.07\textbar0.07\textbar\textbf{0.06} \\
        Mitts & 28.00\textbar27.50\textbar\textbf{28.22} & 0.80\textbar0.77\textbar\textbf{0.82} & 0.25\textbar0.28\textbar\textbf{0.21} \\
        Keyboard & 28.55\textbar28.60\textbar\textbf{28.69} & 0.87\textbar0.88\textbar\textbf{0.89} & 0.09\textbar0.09\textbar\textbf{0.08} \\
        Espresso & 25.72\textbar25.80\textbar\textbf{26.12} & 0.88\textbar0.88\textbar\textbf{0.89} & 0.09\textbar0.09\textbar\textbf{0.08} \\
        \bottomrule
    \end{tabular}}
    
    \label{tab:render_hyper}
\end{table}

\begin{table}[!t]
\caption{Quantitative results of rendered image quality for various scenes in the D-NeRF dataset. Sequential Baseline (Left), Joint Baseline (Middle), Ours (Right). Our method outperforms the baselines on synthetic dataset.}
    \centering
    \resizebox{\linewidth}{!}{
    \begin{tabular}{lccc}
        \toprule
        \textbf{Scene}& \textbf{PSNR(dB)↑} & \textbf{SSIM↑} & \textbf{LPIPS↓} \\
        \midrule
        Standup & 37.24\textbar37.82\textbar\textbf{38.11} &
        0.99\textbar0.98\textbar\textbf{0.98} & 
        0.01\textbar0.009\textbar\textbf{0.007} \\
        Jumping & 34.81\textbar35.23\textbar\textbf{35.53} & 0.97\textbar0.98\textbar\textbf{0.98} & 0.02\textbar0.01\textbar\textbf{0.01} \\
        Hook & 32.60\textbar32.49\textbar\textbf{32.78} & \textbf{0.97}\textbar0.96\textbar\textbf{0.97} & 0.02\textbar0.02\textbar\textbf{0.01} \\
        Trex & 33.41\textbar33.89\textbar\textbf{34.62} & 0.97\textbar0.98\textbar\textbf{0.98} & 0.02\textbar0.01\textbar\textbf{0.01} \\
        \bottomrule
    \end{tabular}}
    
    \label{tab:render_dnerf}
\end{table}

\section{More Quantitative Results} 
\label{supp_sec2}
We provide the rendering metrics results of our method and baselines for different scenes on the HyperNeRF dataset and D-NeRF dataset in Tab.~\ref{tab:render_hyper} and Tab.~\ref{tab:render_dnerf}. We show the segmentation metrics results in Tab.~\ref{tab:segment_supp}. The experimental results demonstrate the effectiveness of our method.

\section{Comparison to DGD}
\label{supp_sec3}
\begin{figure*}[!htpb]
    \centering    \includegraphics[width=1\linewidth]{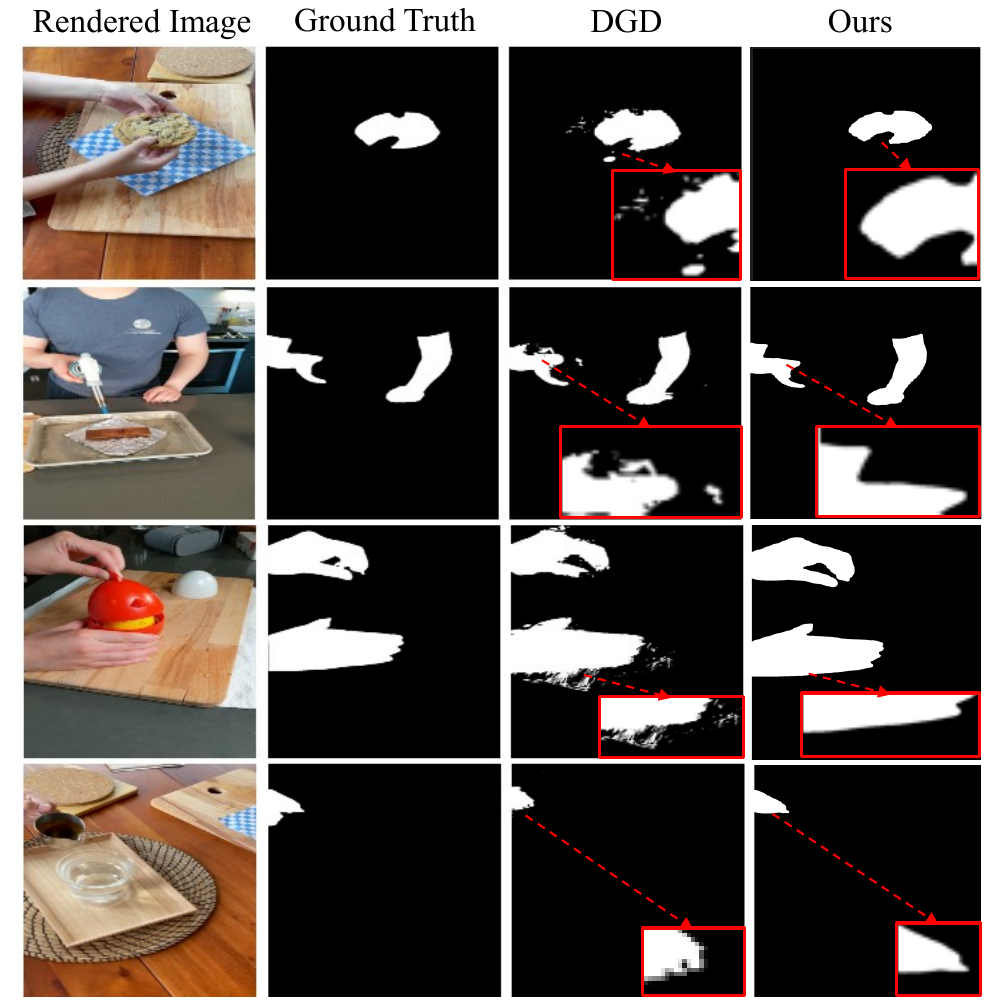}
    \caption{More visual comparisons of our method and DGD. Our method significantly reduces artifacts and noise.}
    \label{duibi_DGD}
\end{figure*}

In Tab.~\ref{tab:segment} and Fig.~\ref{result} (b) of the main text, we show the comparison results of our method and DGD method in segmentation accuracy. We show more comparison results in Tab.~\ref{tab:segment_supp} and Fig.~\ref{duibi_DGD}. DGD is prone to generate noise at the boundary of dynamic objects in the segmentation process, and our method greatly reduces the artifacts.

\section{Semantic Features at Different Scales}
\label{supp_sec4}
\begin{figure*}[!t]
    \centering    
    \includegraphics[width=0.8\linewidth]{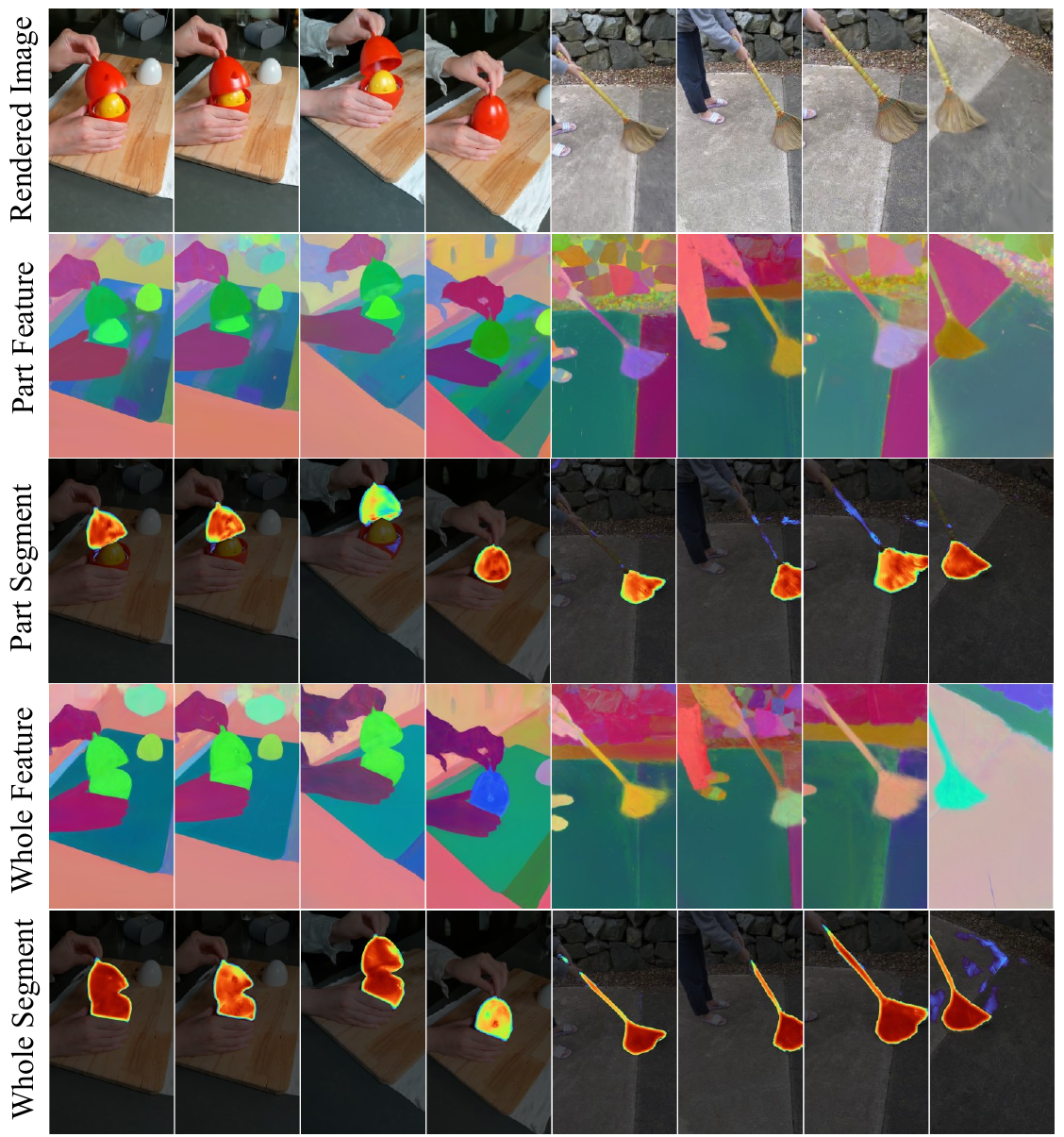}
    \caption{Visualization results of multi-scale semantic features and segmentation (Broom, Chickchicken).}
    \label{scales_1}
\end{figure*}

\begin{figure*}[!htpb]
    \centering    
    \includegraphics[width=0.85\linewidth]{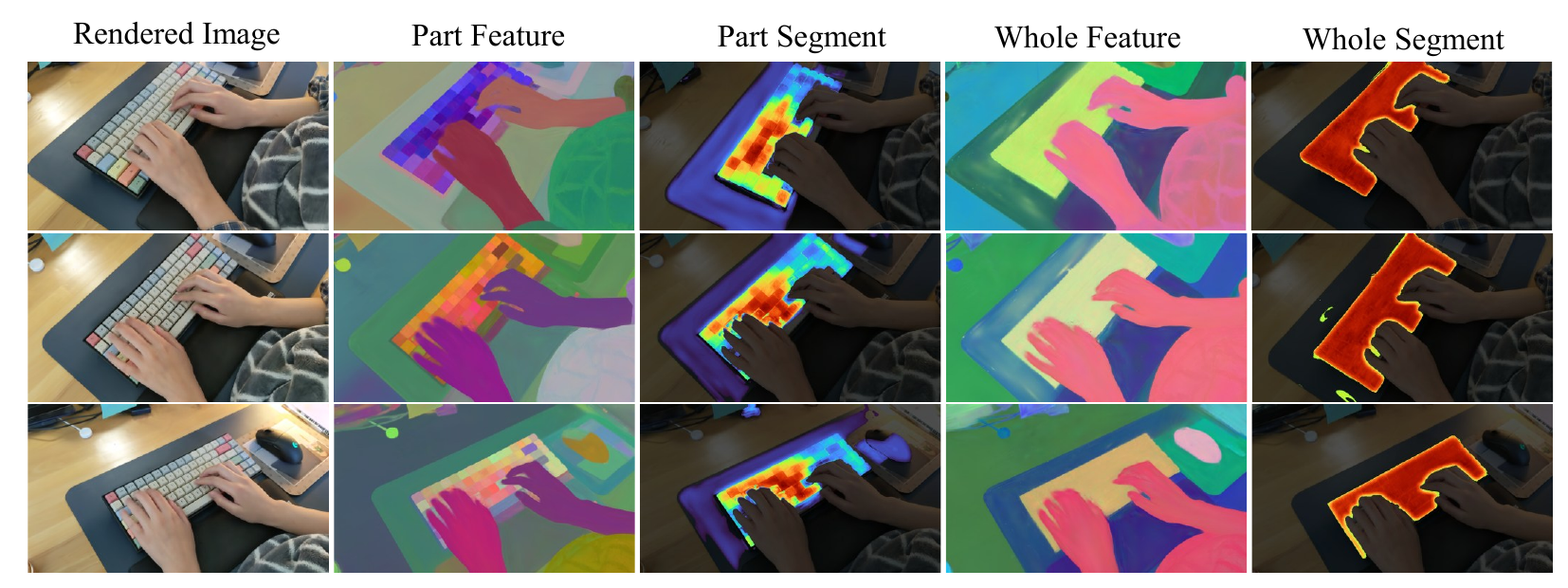}
    \caption{Visualization results of multi-scale semantic features and segmentation (Keyboard).}
    \label{scales_2}
\end{figure*}

Our method can produce semantic feature and segmentation results at different scales. We present visualizations in Fig.~\ref{scales_1} and Fig.~\ref{scales_2}, where we demonstrate multi-scale segmentation of broom, toys, and keyboard. Compared to Fig.~\ref{result} (c) in the text, SA4D cannot segment internal small objects, our method can perceive objects at different scales.

\section{More Visualization Results}
\label{supp_sec5}
\begin{figure*}[!htpb]
    \centering    
\includegraphics[width=1\linewidth]{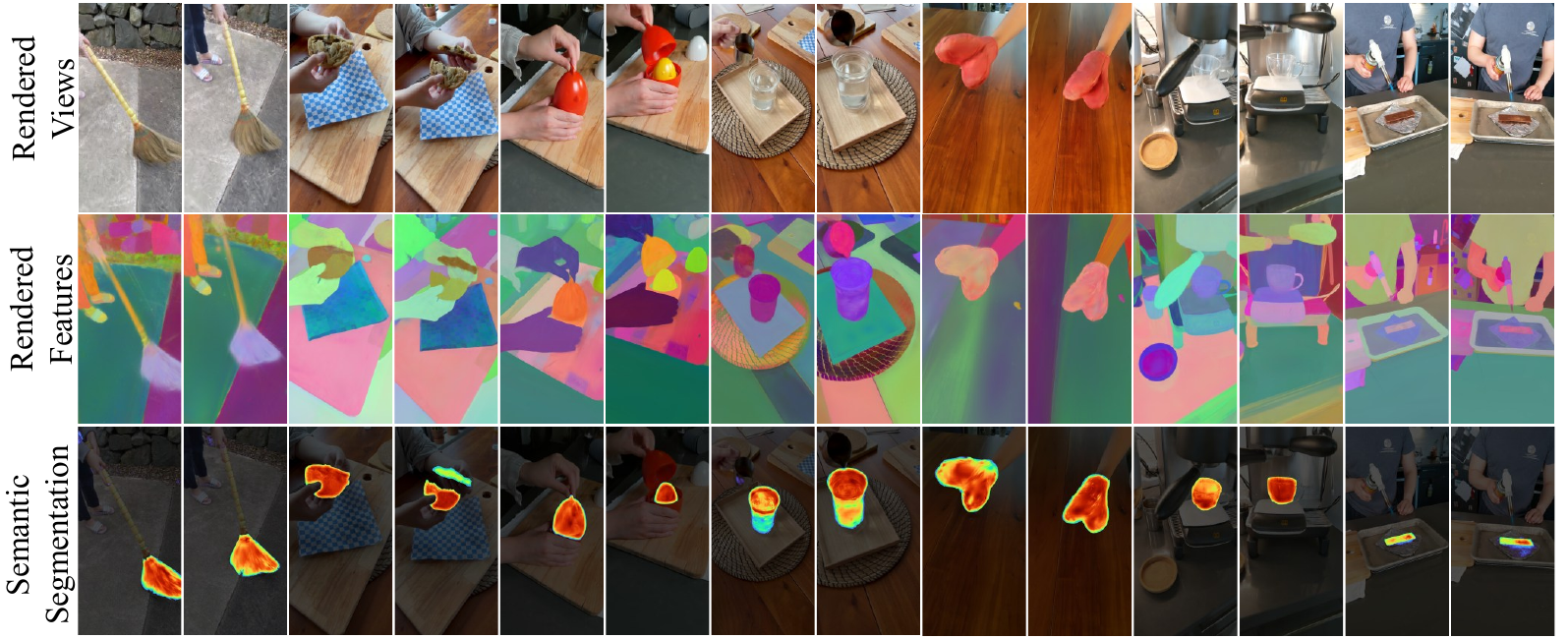}
    \caption{More visualization results on the HyperNeRF dataset (960×536).}
    \label{renders_hyper}
\end{figure*}

\begin{figure*}[!htpb]
    \centering    
\includegraphics[width=1\linewidth]{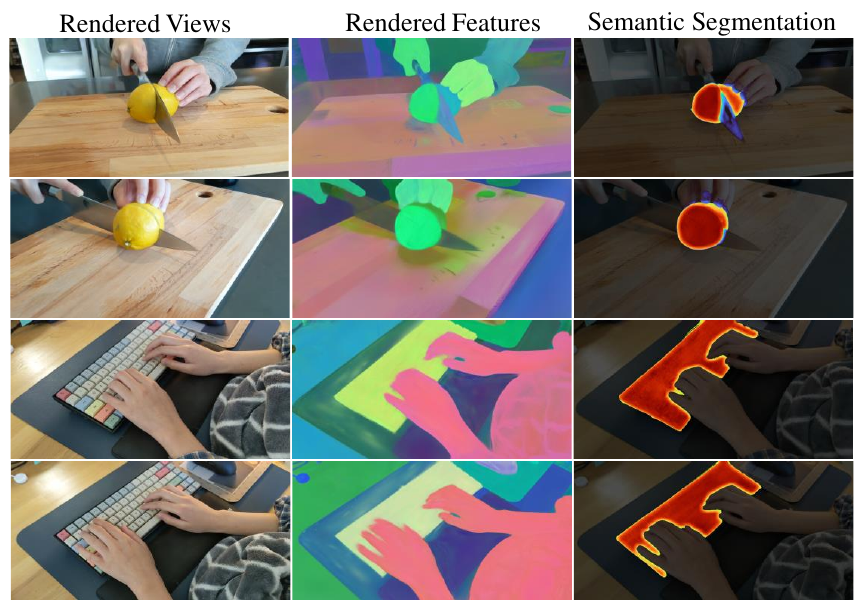}
    \caption{More visualization results on the HyperNeRF dataset (536×960).}
    \label{renders_lemon}
\end{figure*}

\begin{figure*}[!htpb]
    \centering    
    \includegraphics[width=1\linewidth]{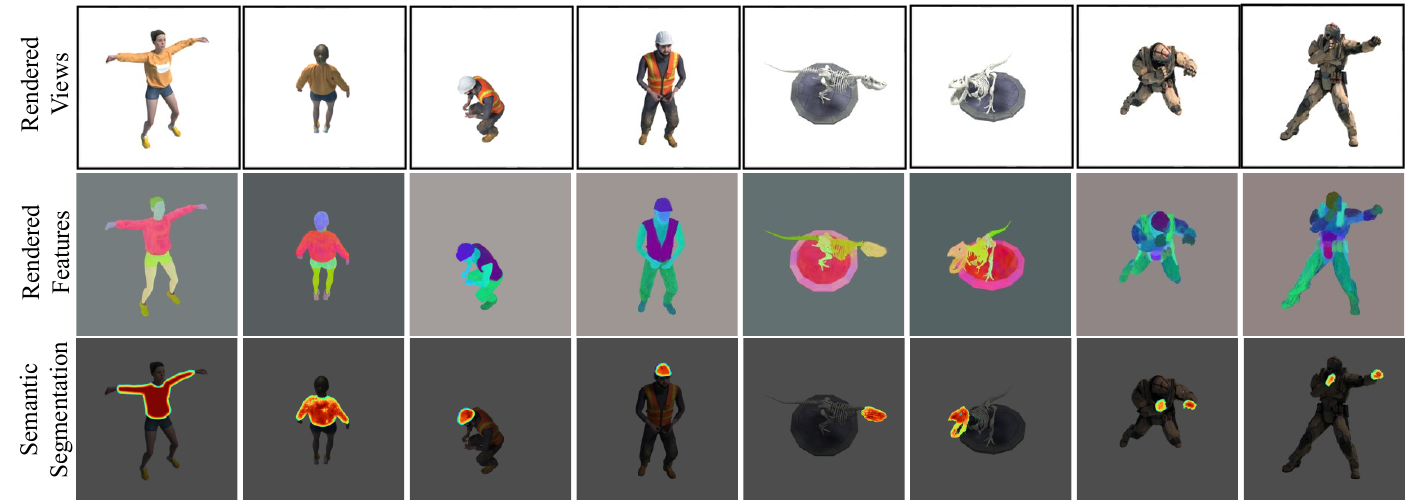}
    \caption{More visualization results on the D-NeRF dataset (800×800).}
    \label{renders_dnerf}
\end{figure*}

In Fig.~\ref{renders_hyper}, Fig.~\ref{renders_lemon}, Fig.~\ref{renders_dnerf} we show the visualizations of our approach on both the HyperNeRF and D-NeRF datasets, with the integrated output of high-quality novel views rendering, semantic feature, and segmentation.

\begin{table}[!t]
\caption{mIoU for object segmentation from the HyperNeRF (top six) and D-NeRF (bottom three). Due to the unavailability of DGD's annotation and segmentation code, we can only use the results presented in their paper. Our method achieves higher segmentation accuracy. We choose the same segmentation object as DGD to ensure fairness.}
    \centering
    \resizebox{\linewidth}{!}{
    \begin{tabular}{lcccc}
        \toprule
        \textbf{Scene (Prompt)} & \textbf{Sequential} & \textbf{Joint} & \textbf{DGD$^*$} & \textbf{Ours} \\ 
        \midrule
        Broom (Broom) & 44.32 & 64.25 & -- & \textbf{78.46}\\
        Cookie (Cookie) & 84.64 & 88.59 & 78.65 & \textbf{92.19}\\
        Lemon (Lemon) & 73.93 & 84.02 & -- & \textbf{90.89}\\
        Chicken (Hands) & 74.53 & 81.03 & 77.92 & \textbf{90.84}\\
        Chocolate (Hands) & 68.49 & 74.75 & 75.98 & \textbf{80.14}\\
        Americano (Hands) & 77.38 & 88.29 & 78.98 & \textbf{91.56}\\
        \midrule
        Standup (Helmet) & 89.21 & 90.08 & 93.80 & \textbf{95.66}\\
        Jumping (Hands) & 31.25 & 75.69 & 68.35 & \textbf{83.79}\\
        Trex (Skull) & 70.87 & 85.66 & 70.39 & \textbf{90.67}\\
        \bottomrule
    \end{tabular}}    
    \label{tab:segment_supp}
\end{table}

\begin{table}[!htpb]
\caption{Our average rendering speed and storage for scenes from the HyperNeRF dataset (top four) and D - NeRF dataset (bottom two). Our method achieves a good balance between storage efficiency and rendering speed.}
\centering
\setlength{\tabcolsep}{18pt} {
\begin{tabular}{l|cc}
\toprule
\textbf{Dataset} & \textbf{FPS} & \textbf{Storage(MB)} \\
\midrule
Broom ($\bar{D}$=0.21) & 25 & 67 \\
Americano ($\bar{D}$=0.07) & 28 & 65 \\
Cookie ($\bar{D}$=0.05) & 30 & 55 \\
Lemon ($\bar{D}$=0.03) & 29 & 47 \\
\midrule
Hook ($\bar{D}$=0.01) & 115 & 25 \\
Standup ($\bar{D}$=0.007) & 118 & 24 \\
\bottomrule
\end{tabular}}
\label{tab:model_performance}
\end{table}

\section{Mask Annotation}
\label{supp_sec6}
\begin{figure*}[!htpb]
    \centering    \includegraphics[width=1\linewidth]{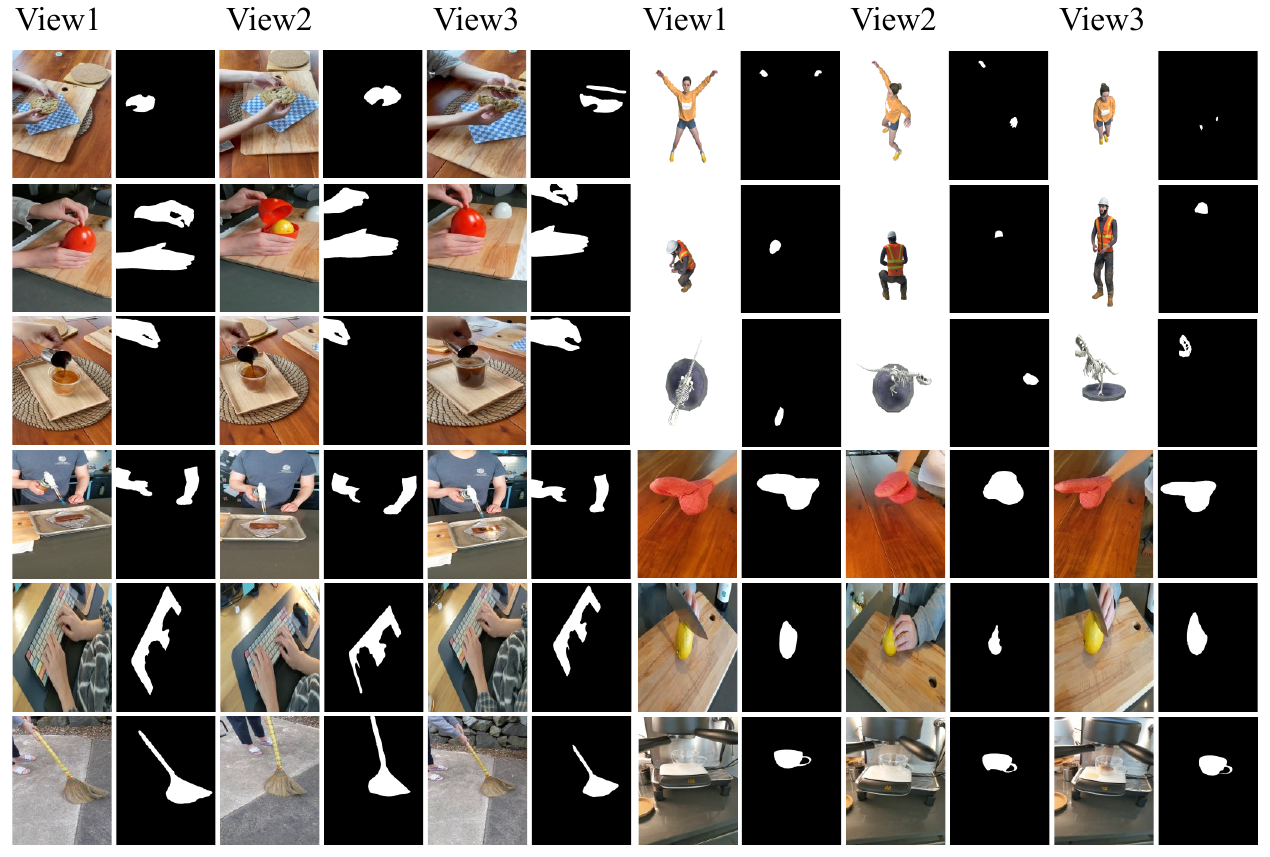}
    \caption{Visualizations of mask annotations on HyperNeRF and D-NeRF dataset.}
    \label{annotation}
\end{figure*}

Due to the lack of segmentation instance ground truth in the HyperNeRF and D-NeRF datasets, and the unavailability of annotated files from DGD, we attempt manual annotation of these datasets. We annotate 9 scenes from HyperNeRF and 4 scenes from D-NeRF. We utilize SAM for initial segmentation, followed by manual refinement to obtain the more accurate mask. Examples of annotations are provided in Fig.~\ref{annotation} . The annotation ground truth will be publicly available.

\section{Speed and Memory Requirements}
\label{supp_sec7}
In Tab.~\ref{tab:model_performance}, we show our rendering speed and storage. Compared to DGD, our proposed approach requires significantly less memory. While DGD training typically necessitates a 40GB A100 GPU, our method can be successfully trained on an RTX 3090 with only 8GB of memory. Our method exhibits reduced resource consumption while achieving improved model performance.

\end{document}